\documentclass[preprint,12pt]{elsarticle}

\usepackage{amssymb}
\usepackage{amsmath}
\newcommand{\figref}[1]{Fig.~\ref{#1}}
\usepackage{caption}
\usepackage{makecell}
\usepackage{orcidlink}
\usepackage{tabularx} 
\usepackage{arydshln}
\usepackage{algpseudocode}
\usepackage{graphicx}
\usepackage{bigstrut,multirow,rotating,booktabs}
\usepackage{cleveref}

\Crefname{figure}{Fig}{Figures}
\biboptions{sort&compress}

\begin{document}

\begin{frontmatter}



\title{Incremental Multiview Point Cloud Registration}


\author{Xiaoya Cheng} 
\author{Yu Liu} 
\author{Maojun Zhang} 
\author{Shen Yan\corref{cor1}}
\cortext[cor1]{Corresponding author at: College of Systems Engineering, National University of Defense Technology, Changsha, Hunan 410000, China.}
\begin{abstract}
  In this paper, we present a novel approach for multiview point cloud registration. Different from previous researches that typically employ a global scheme for multiview registration, we propose to adopt an \textit{incremental} pipeline to progressively align scans into a canonical coordinate system. Specifically, drawing inspiration from image-based 3D reconstruction, our approach first builds a sparse scan graph with scan retrieval and geometric verification. Then, we perform incremental registration via initialization, next scan selection and registration, Track create and continue, and Bundle Adjustment. Additionally, for detector-free matchers, we incorporate a Track refinement process. This process primarily constructs a coarse multiview registration and refines the model by adjusting the positions of the keypoints on the Track. Experiments demonstrate that the proposed framework outperforms existing multiview registration methods on three benchmark datasets. The code is available at https://github.com/Choyaa/IncreMVR.
\end{abstract}


\end{frontmatter}



\section{Introduction}
\label{sec:intro}
\begin{figure}[t]
   \centering
   \includegraphics[width=0.95\linewidth]{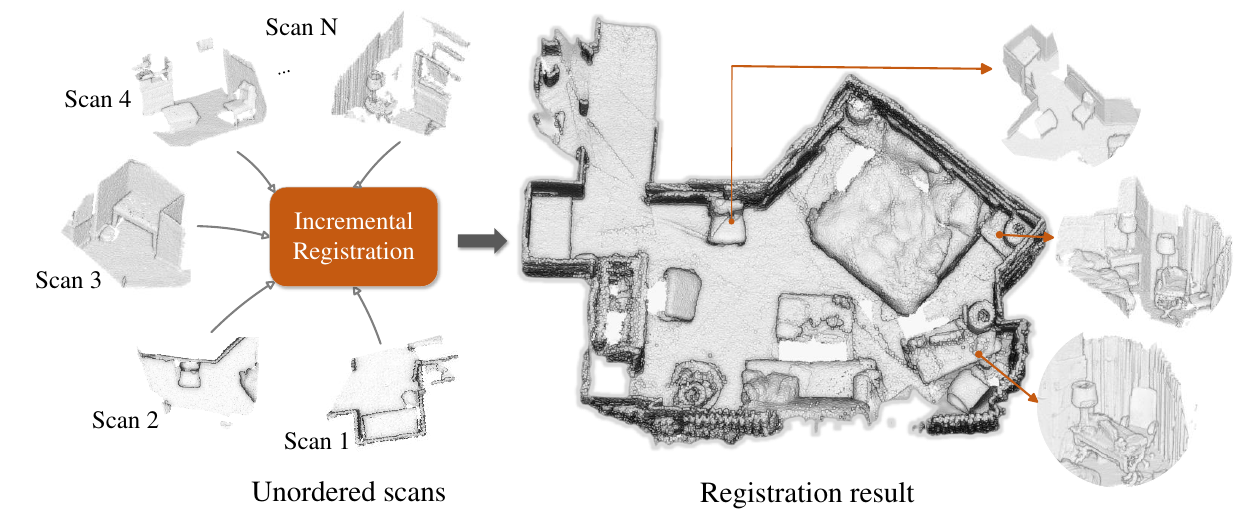}
   \caption{\textbf{Results of our incremental registration.} The input consists of multiple unordered scans captured from different viewpoints, while the output is a single aligned result representing the reconstructed scene in a uniform coordinate system. Please zoom in to see sharp geometry details in the rightmost areas.}
   \label{fig:intro}
\end{figure}
Point cloud registration serves as a fundamental prerequisite for many downstream tasks in 3D vision, encompassing 3D object detection~\cite{xiang2014beyond}, 3D segmentation~\cite{nguyen20133d} and 3D reconstruction~\cite{dong2020registration,guo2020deep,huang2021comprehensive}. Most recent registration methodologies~\cite{ao2021spinnet,bai2020d3feat,gojcic2019perfect,huang2021predator,qin2022geometric,wang2022you,choy2019fully} mainly concentrate on pairwise registration of two partial point clouds, which can only recover a segment of the scene. To achieve a completed scene reconstruction, all point clouds (scans) should be registered into a uniform coordinate reference system, a process known as \textit{multiview registration}. Due to its complexity, multiview point cloud registration has received limited attention, with only few recent studies proposing solutions~\cite{wang2023robust,huang2019learning,yew2021learning,yang2016automatic,gojcic2020learning,dong2018hierarchical}.

Recent progress in this field has been largely driven by \textit{global} multiview registration methods~\cite{gojcic2020learning,yew2021learning,huang2019learning,wang2023robust}. These typically employ a two-step workflow. The first step recovers the relative transformation between two partial point clouds, usually using feature matching (detector-based~\cite{wang2022you,choy2019fully,ao2021spinnet} or detector-free~\cite{qin2022geometric}), in conjunction with a robust estimator such as RANSAC. The second step jointly optimizes the relative transformations from the first step to recover the absolute point cloud poses, which is called Transformation Synchronization. To prune noisy or even incorrect relative transformations between two point clouds, cycle consistency~\cite{huang2019learning} and Iterative Reweighting Least Square (IRLS)~\cite{gojcic2020learning,yew2021learning,wang2023robust} are often employed in the second step.

Contrary to the \textit{global} scheme of multiview point cloud registration, the reconstruction from unordered photo collections is still predominantly governed by the \textit{incremental} Structure-from-Motion (SfM)~\cite{schonberger2016structure,he2023detector,wu2013towards} strategy. This process typically builds a view graph by feature extraction and matching, followed by geometric verification. This graph serves as the basis for the reconstruction stage, which seeds the model with a carefully selected two-view reconstruction, then incrementally registering new images, triangulating scene points, filtering outliers, and refining the reconstruction using Bundle Adjustment (BA). Numerous well-established methods~\cite{schonberger2016structure,he2023detector,agarwal2011building,frahm2010building,snavely2006photo,wu2013towards}, open-source systems such as Bundler~\cite{snavely2008bundler} and COLMAP~\cite{schonberger2016structure}, and commercial software~\cite{contextcapture,metashape} exist that can accurately and robustly recover large-scale scenes.

Motivated by the success of image-based 3D reconstruction, we introduce a novel \textit{incremental} framework tailored for multiview point cloud registration. Given a set of potentially overlapping scans as input, our method outputs a globally consistent pose for each of the input scan, as depicted in \figref{fig:intro}. 
Specifically, our method starts with feature matching between pairs of point clouds, thereby forming a scan graph where each node signifies a scan and each edge denotes the matching relationship between scans. The point cloud retrieval technique (with a global feature) and geometric verification (e.g., the number and rate of inliers) are used to eliminate mismatches, resulting in a sparse scan graph with fewer, yet more reliable edges.
We then select initial pairs based on this scan graph, considering both the reliability of the match and its position in the graph. Subsequently, we adopt an incremental strategy to select and register the next scan into a predefined canonical coordinate system. This registration involves establishing 3D-3D correspondences between keypoints in the newly selected scan and 3D landmarks from previously registered scans.
To mitigate error accumulation during the incremental process, we employ a combination of local Bundle Adjustment (which optimizes scan poses only) and global Bundle Adjustment (which optimizes both scan poses and reconstructed 3D landmarks).
Notably, our framework exhibits flexibility by accommodating a detector-free matching scheme with minimal adjustments. In this configuration, we initiate the process by constructing an initial multiview point cloud registration model based on quantified matches. After that, we iteratively refine this model, leveraging track scores during fine matching to achieve higher accuracy.


We conduct extensive experiments to validate the efficacy of our proposed method on 3D(Lo)Match~\cite{huang2021predator,zeng20173dmatch}, ScanNet~\cite{dai2017scannet}, and ETH~\cite{pomerleau2012challenging} datasets. The results indicate that our method outperforms other comparative methods on the 3D(Lo)Match and ScanNet datasets. On the ETH dataset, our method achieves results comparable to the state-of-the-art global methods, with a registration recall rate nearing 99\%.

In conclusion, our primary contributions are as follows:
\begin{itemize}
    \item A novel incremental framework for multiview point cloud registration. 
    \item Supporting detector-based and detector-free registration matchers.
    \item Demonstrating the effectiveness of our approach on three benchmarks.
\end{itemize}


\section{Related Work}
\label{sec:related}
\subsection{Pariwise Registration}
Traditional pairwise registration methods~\cite{besl1992method, AndreaCensi2008AnIV, forstner2017efficient, iglesias2020global,liang2020precise, Choy_Park_Koltun_2019} typically frame point cloud registration as an energy minimization problem, which defines a distance function and provides a closed-form solution, such as ICP(Iterative Closest Point)~\cite{besl1992method} and its variants~\cite{AndreaCensi2008AnIV, SrikumarRamalingam2013ATO}.

For learning-based methods, a primary category of current pairwise registration algorithms are detector-based methods~\cite{bai2020d3feat, wang2022you,gojcic2019perfect,Huang_Gojcic_Usvyatsov_Wieser_Schindler_2021}. Theses methods rely on finding and matching repeatable keypoints~\cite{bai2020d3feat, Huang_Gojcic_Usvyatsov_Wieser_Schindler_2021}, followed by the estimation of the transformation using a robust estimator, such as RANSAC~\cite{Fischler_Bolles_1981}. To improve the matching accuracy, some algorithms incorporate outlier removal techniques~\cite{huang2021predator,Bai_Luo_Zhou_Chen_Li_Hu_Fu_Tai_2021} to prune incorrect correspondences. Recently, the advent of detector-free methods~\cite{Yu_Li_Saleh_Busam_Ilic_2021, qin2022geometric}, which directly establish dense matching between point clouds without the reliance on repeatable keypoints, have been proposed. These methods have demonstrated superior performance, particularly in environments devoid of detailed structure.
\subsection{Multi-view  Registration}
Multiview registration aims to align point clouds obtained from different viewpoints of the same object or scene. This task is inherently complex due to several factors. First, real-scanned point clouds frequently exhibit incompleteness, posing a significant challenge for accurate alignment. Second, even minor errors in partial registration can propagate and ultimately result in complete failure in the final reconstructed point cloud.


A majority of the established methods have adopted a \textit{global} approaches~\cite{Huang_Liang_Bajaj_Huang_2017, Arrigoni_Rossi_Fusiello_2016,huang2019learning,Chatterjee_Govindu_2018, Lee_Civera} to solve this problem. These approaches initially calculate the relative transformations between pairs of point clouds, followed by the computation of the absolute poses of the point clouds through graph optimization. To mitigate the impact of erroneous pairwise transformations, recent studies have proposed the construction of the reliable pose graph~\cite{Bhattacharya_Govindu_2019, Huber_Hebert_2003, Yew_Lee, huang2019learning} or the utilization of Iteratively Reweighted Least Square (IRLS)~\cite{huang2019learning, Bernard_Thunberg_Gemmar_Hertel_Husch_Goncalves_2015, Huang_Liang_Bajaj_Huang_2017}. 

Besides, several straightforward \textit{incremental} multiview point cloud registration algorithms~\cite{Evangelidis_Horaud_2018, Guo_Sohel_Bennamoun_Wan_Lu_2014, DONG201861, WU202365, Zhu_Zhu_Li_Li_Cui_2016} have been proposed. These algorithms~\cite{ Zhu_Zhu_Li_Li_Cui_2016, Huber_Hebert_2003, Kelbe_aar_paul} typically begin by constructing a Minimum Spanning Tree (MST) and then progressively register point clouds to a coordinate system, moving from the root node to the leaf nodes. Hierarchical methods~\cite{DONG201861, ZHU2017477, WU202365} extend this strategy by integrating multiple groups of point clouds into the evolving registered model. However, these methods are highly sensitive to the construction of the MST. A single error during MST construction can lead to catastrophic failure in the overall multiview point cloud registration process. Moreover, as the registration progresses, accumulated errors gradually escalate.

To mitigate these challenges, we draw inspiration from recent advancements in image-based 3D reconstruction. Instead of relying solely on an MST, we propose constructing a more intricate scan graph. This graph captures the tentative information provided by multiple overlaps, allowing us to incrementally register subsequent scans to the initial model. Additionally, we leverage Bundle Adjustment optimization techniques to mitigate accumulated errors, facilitating the construction of large-scale scenes.
\begin{figure}[t]
   \centering
   \includegraphics[width=0.95\linewidth]{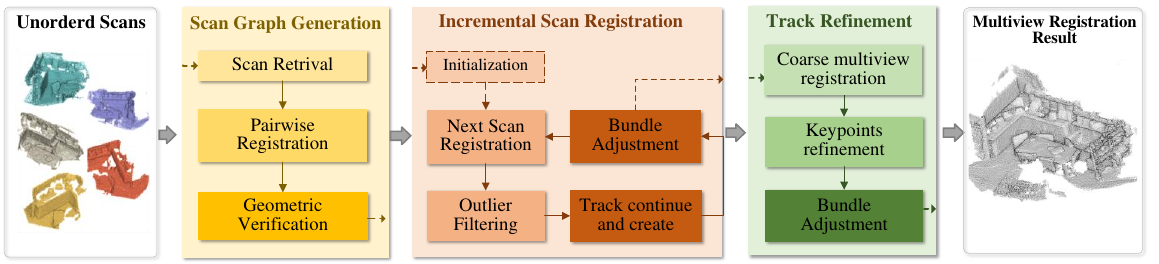}
   \caption{\textbf{Overview of the proposed method.} Our method has three components: scan graph generation (\cref{sec:scene_graph}), incremental scan registration (\cref{sec:scan_reg}) and track refinement (\cref{sec:track_ref}, optionally for detector-free matchers).}
   \label{fig:main}
\end{figure}

\section{Method}
\label{sec:method}
An overview of our incremental approach is exhibited in \figref{fig:main}. Given an unordered collection of $N_s$ scans $\mathcal{P} =\left\{ P_i|i\,\,=\,\,0,...,N_s-1 \right\}$, our objective is to recover the global poses $\mathcal{T} =\left\{ T_i = (R_i,t_i) \in SE(3) |i\,\,=\,\,0,...,N_s-1 \right\}$. To achieve this, our first step involves the construction of a sparse scan graph (Section~\ref{sec:scene_graph}). Then, we initiate the multiview registration process by carefully selecting a two-scan seeding and incrementally register new scans to construct a complete scene (Section~\ref{sec:scan_reg}). Finally, we extend our method to incorporate a detector-free matching scheme, which is founded on a proposed track refinement process (Section~\ref{sec:track_ref}).
\begin{figure}[t]
   \centering
   \includegraphics[width=0.95\linewidth]{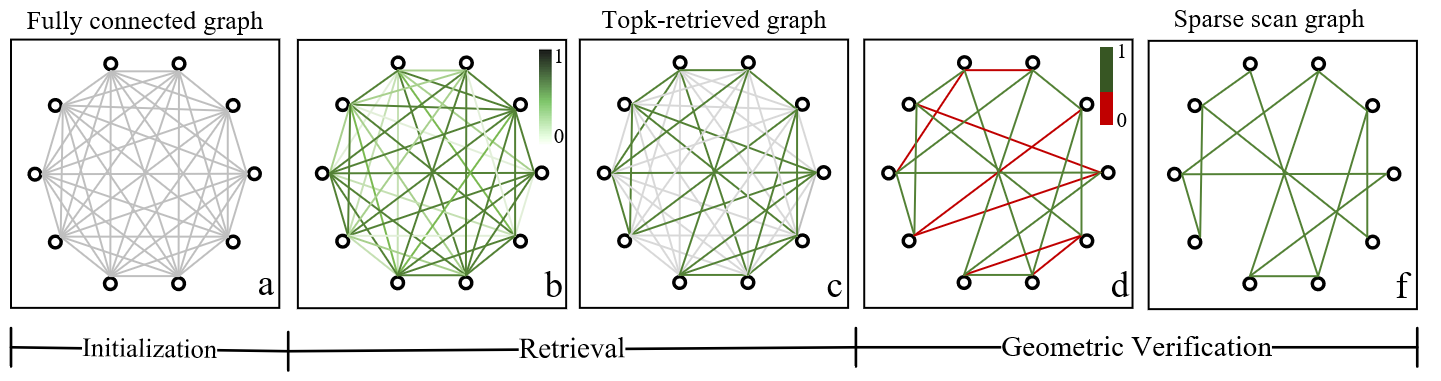}
   \caption{\textbf{The process of sparse scan graph generation.} The sparse scan graph is established by leveraging both scan retrieval and geometric verification.}
   \label{fig:graph_generation}
\end{figure}
\subsection{Sparse Scan Graph Construction}
\label{sec:scene_graph}
This process aims to construct a sparse scan graph for multiview registration. We formulate this scan graph as $\mathcal{G}(\mathcal{V}, \mathcal{E})$, where each node $v_i \in \mathcal{V}$ corresponds to a scan $P_i$ associated with its extracted keypoints $\{ p_a \}$ and each edge $e_{ij} \in \mathcal{E}$ signifies the results of keypoints matching $\mathcal{M}_{ab} = \{ (p_a, p_b) \}$ and the relative transformation $(R_{ij},t_{ij})$. 

Our approach first calculates an overlap score $s_{ij}$ for each scan pair $(P_i, P_j)$, employing advanced deep retrieval techniques. We select the top $k$ scan pairs with the highest overlap scores to establish the coarse edges $\mathcal{E}_{coar}$. While the retrieval method effectively sparsifies the scan graph, its global feature representation tends to overlook point-level details, leading to the inclusion of some erroneous edges. 
To address this limitation, we introduce a strategic geometric verification approach that leverages local feature information, thereby eliminating the erroneous edges. In detail, we employ a point cloud feature extractor to obtain keypoints $\{ p_a \}$ for each node $v_i$. Subsequently, a feature matching algorithm establishes 3D-3D matching relationships $\mathcal{M}_{ab}$ for each edge in $\mathcal{E}_{coar}$. 
The SVD RANSAC algorithm is then engaged to estimate the 6-DoF transformation $(R_{ij}, t_{ij})$, concurrently providing inlier count and proportion of the matching. By setting a threshold for the number $\tau_{n}$ and proportion $\tau_{p}$ of inliers, we further filter out incorrect edges, resulting in a reliable sparse scan graph $\mathcal{G} \left( \mathcal{V}, \mathcal{E} \right) $. The whole generation process is illuminated in \figref{fig:graph_generation}.

\begin{figure}[ht]
  \centering
  \includegraphics[width=\textwidth]{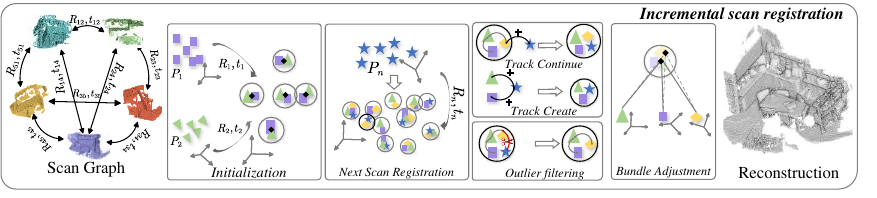}
  \caption{\textbf{Incremental Scan Registration.} Given a scan graph with relative transformations, the incremental process first registers the initial pair into the global coordinate system and creates initial tracks. Then, it iteratively registers the next scan and manages the track data through continue, create, and filtering operations. The Bundle Adjustment is performed to optimize tracks throughout the entire process.}
  \label{fig:incremental_registration}
\end{figure}

\subsection{Incremental Scan Registration}
\label{sec:scan_reg}
This process is designed to align all scans within a canonical coordinate system, leveraging the relational constraints within a sparse scan graph to determine the absolute pose of the scan. The whole process builds upon the fundamental concept of incremental registration, as illustrated in \figref{fig:incremental_registration}.



\paragraph{\textbf{Initialization}} 
Selecting an appropriate initial pair (namely, a certain edge in scan graph $\mathcal{G}$) is critical, since the multiview registration may never recover from a poor choice. Our approach takes into account two principal criteria: the quantity of matches and the centrality of nodes within the scan graph. The former is essential to prevent registration failure at the beginning, while the latter ensures that the process originates from the center of the scan graph, thereby reducing error accumulation. 

To realize this, we devise a simple yet effective algorithm. Initially, we sort the set of scans $\mathcal{V}$ by closeness centrality, and successively select these scans in a descent manner. For a selected scan $v_i$, we identify an edge $e_{ij}$ who characters the maximum matching number $m$ among all neighboring edges enclosing scan $v_i$. If the matching number $m$ exceeds a predefined threshold $\tau_{m}$, we terminate the iteration and determine this edge as initialization. 

\begin{figure}[t]
   \centering
   \includegraphics[width=0.95\linewidth]{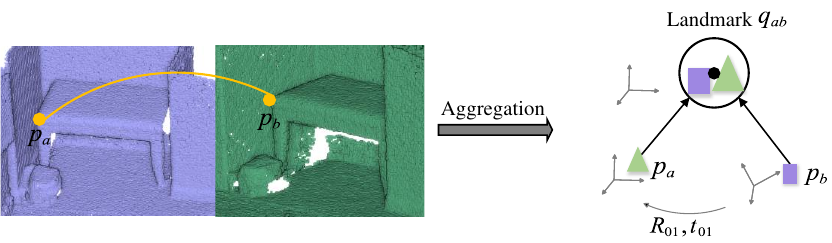}
   \caption{\textbf{Aggregation.} Two sets of point cloud frames (colored by purple and green), share a common scanning area. Corresponding keypoints from each set are merged into a landmark point in the global coordinate system through an aggregation operation.}
   \label{fig:aggregation}
\end{figure}

Let us consider the initial scan pairs selected as $P_0$ and $P_1$, encompassing corresponding keypoints denoted as $p_a \in P_0$ and $p_b \in P_1$. Given their relative transformation $(R_{01},t_{01})$ retrieved from the scan graph $\mathcal{G}$, as depicted in \figref{fig:aggregation}, we assign the global pose $(R_0,t_0) = (\mathbf{I},\mathbf{0})$ to the scan $P_0$, and $(R_1,t_1) = (R_{01},t_{01})$ to scan $P_1$.
Subsequently, based on the established matches $\mathcal{M}_{ab}$ and the global poses $(R_0,t_0), (R_1,t_1)$, we engage in the generation of 3D landmarks $\{ q_{ab} \}$ through an operation termed \textit{aggregation}. This process mirrors the \textit{triangulation} technique utilized in image-based 3D reconstruction. The 3D landmarks $\{ q_{ab} \}$ are derived by resolving multiple linear equations, each dedicated to minimizing the re-projection error subsequent to point cloud transformation, as expressed by:
\begin{equation}
    \begin{aligned}
    \label{equ:reproj}
    & p_a - (R_0 \cdot q_{ab} + t_0) = 0, \\
    & p_b - (R_1 \cdot q_{ab} + t_1) = 0. \\
    \end{aligned}
\end{equation}
We employ a data structure termed Track to record the generated results, where each Track is represented by 3D landmarks and the keypoints that generate them $\{ \mathbb{T}_j \} = \{ q_{ab}, \{ p_a, p_b \} \}$.

\paragraph{\textbf{Next Scan Selection}}
After the initialization, our objective is to identify a scan, denoted as $P_l$ that has the maximum visibility of the current 3D landmarks for the subsequent registration process. To achieve this, we conduct an exhaustive search across the entire collection of unregistered scans.




For each candidate scan $P_c$, we compute the sum of tracks visible to the keypoints $\{ p_c \} \in P_c$. The scan that exhibits the highest count of observable tracks (denoted as $m$) is designated as the subsequent scan $P_l$, provided that $m$ surpasses a predefined threshold $\tau_l$. Failure to meet this threshold prompts termination of the incremental registration process, necessitating a re-initiation within the graph of unregistered scans.
\begin{figure}[t]
   \centering
   \includegraphics[width=0.95\linewidth]{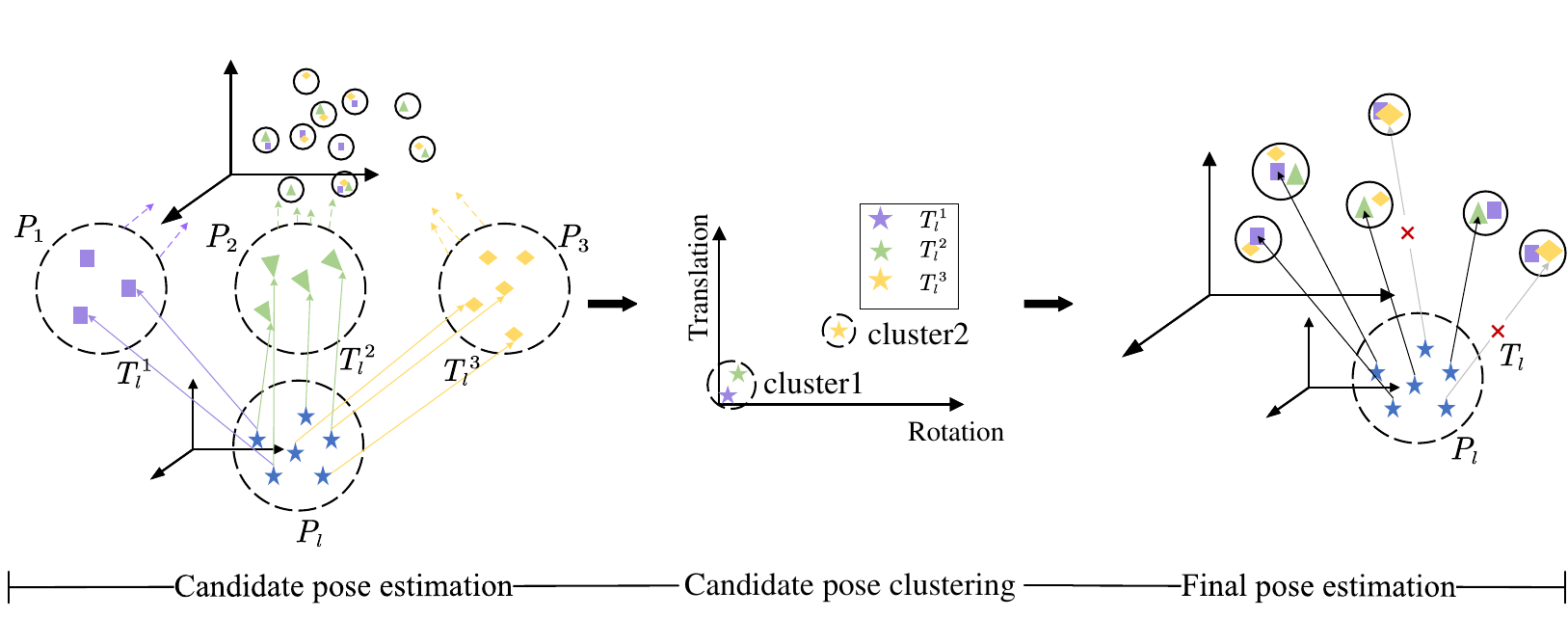}
   \caption{\textbf{Next scan registration.} We first generate several pose candidates leveraging keypoints-to-landmarks matches across neighboring scans. Then, we cluster these candidate poses,  retaining the most largest cluster and amalgamating their corresponding 3D-3D matches. Finally, we solve for a robust pose within a RANSAC loop.}
   \label{fig:cluster}
\end{figure}

\paragraph{\textbf{Next scan registration}}
The determination of the absolute pose of the next scan is typically achieved by directly solving the transformation using keypoint-to-landmark correspondences. However, a significant incidence of mismatches has been observed within these correspondences, leading to a notable deterioration in the registration quality of the next scan. our efforts to enhance the results using robust algorithms, including RANSAC, we discovered that over 20\% of the estimates had a rotation error exceeding $5^{\circ}$.   



To address this issue, we design a clustering algorithm to solve for the next scan pose, as illustrated in ~\figref{fig:cluster}. For the next scan $P_l$, we first establish matches between 3D keypoints and landmarks, denoted as $\left( \mathcal{M}_l^1,\mathcal{M}_l^2,...,\mathcal{M}_l^m \right) $ for neighboring registered scans, identified as $\left( P_1, P_2,..., P_m \right)$. We then independently determine a set of candidate poses $T^{c}_l=\left( T^1_l,T^2_l,...,T^m_l \right) $. we hypothesize that accurate poses are likely to be consistent with one another and cluster together, while the erroneous poses are dispersed. By clustering the candidate poses $T^{c}_l$, we identify the largest cluster and amalgamate their corresponding 3D-3D matches to compute a final pose $T_l = (R_l, t_l)$ within a RANSAC loop. The experimental results validate that the application of the clustering approach can effectively eliminate some noisy 3D-3D matches. 

\begin{figure}[t]
   \centering
   \includegraphics[width=0.95\linewidth]{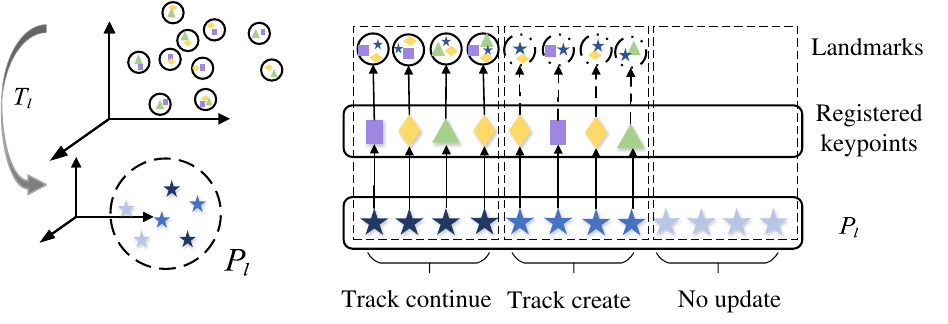}
   \caption{\textbf{Track operation.} The incremental process constructs track constraints at the keypoint level, distinguishing between three categories: those with observable landmarks for track continue, those initiating new tracks without landmarks, and unmatched points excluded from updates.}
   \label{fig:track}
\end{figure}

\paragraph{\textbf{Track continue and create}}
Upon the next scan registration, we manage the Track data through \textit{continue} and \textit{create} operations, as shown in \figref{fig:track}.

For the \textit{continue} operation, given a Track $\mathbb{T}_j$ with 3D landmark $q_{ab}$, and a matching relationship between $q_{ab}$ and $p_c$ in $P_l$, we introduce a new linear equation to aggregate the position of the landmark as $q_{abc}$:
\begin{equation}
    \begin{aligned}
    \label{equ:reproj_2}
    & p_a - (R_0 \cdot q_{abc} + t_0) = 0, \\
    & p_b - (R_1 \cdot q_{abc} + t_1) = 0, \\
    & p_c - (R_l \cdot q_{abc} + t_l) = 0. \\
    \end{aligned}
\end{equation}
In addition, the length of the track is also increased, represented as $\mathbb{T}_j = \{ q_{abc}, \{ p_a,p_b,p_c \} \}$.

For the \textit{create} operation, some keypoints $\{ p_a \}$ in scan $P_0$ or $\{ p_b \}$ in scan $P_1$ may have a matching relationship with some keypoints $\{ p_c \}$ in scan $P_l$. For these points, we employ Equation~\ref{equ:reproj} to aggregate new 3D landmarks, forming new tracks, thereby further expanding the model.

To ensure only reliable keypoints contribute to the track \textit{continue} and \textit{create}, we employ aggregation-RANSAC for robust track update.


\begin{figure}[t]
   \centering
   \includegraphics[width=0.95\linewidth]{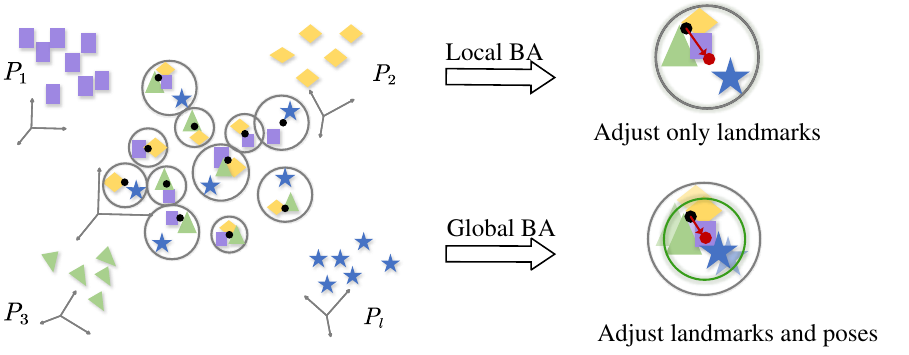}
   \caption{\textbf{Bundle Adjustment.} The upper panel shows local BA, optimizing landmark positions for minimal re-projection errors while keeping poses fixed. The lower panel exhibits global BA, further optimizing poses along with landmarks for precise registration.}
   \label{fig:BA}
\end{figure}
\paragraph{\textbf{Bundle Adjustment}}
To mitigate accumulated errors, we perform Bundle Adjustment (BA) after scan registration and landmark aggregation. Taking both accuracy and efficiency into account, we employ an optimization mechanism that alternates between local and global BA, as depicted in \figref{fig:BA}. 

Specifically, considering a registration state with Tracks $\left\{ \mathbb{T} _j\mid j=1...N \right\}$, where each Track $\mathbb{T}_j$ is associated with a 3D landmark $q_j$ and $M$ keypoints $\{ p_{ij} \mid i=1,..., M \}$ in scans $\{ V_i \}$ with registered poses $(R_i,t_i)$. We formulate an energy function as follows:
\begin{equation}
    \label{equ:ba}
    E=\sum_{i=1}^N{\sum_{j=1}^M{|}}| (R_i \cdot q_j + t_i) - p_{ij} ||_2.
\end{equation}
For local BA, only the scan poses $\{ (R_i,t_i) \}$ are optimized, whereas for global BA, both scan poses $\{ (R_i,t_i) \}$ and aggregated 3D landmarks $\{ q_j \}$ are optimized. This optimization process is implemented by the Ceres solver~\cite{Agarwal_Ceres_Solver_2022}.

\subsection{Track Refinement for detector-free Methods}
\label{sec:track_ref}

\begin{figure}[t]
   \centering
   \includegraphics[width=0.95\linewidth]{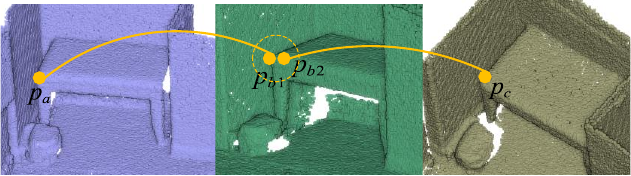}
   \caption{\textbf{Detector-free issue for multiview point cloud registration.} A keypoint $p_b$ is erroneously identified as $p_{b1}$ and $p_{b2}$ in two distinct correspondences, making it impossible to build consistent Track during the multiview point cloud registration process.}
   \label{fig:detect_free}
\end{figure}

\paragraph{\textbf{Detector-free issue}}
Recently, detector-free matching methods, such as GeoTransformer~\cite{qin2022geometric}, perform well in pairwise point cloud registration task. GeoTransformer begins by constructing coarse semi-dense matches between scan pairs (designated as left and right scans), then with all left matches fixed, the right matches are refined to a fine position using high-level point cloud feature. However, due to the lack of fixed keypoints, these methods lead to fragmented Track for multiview point cloud registration, as depicted in \figref{fig:detect_free}. This inconsistency hinders the seamless integration of detector-free matchers into our multi-view registration framework, which relies on fixed keypoints across scans to form consistent Tracks. To solve this problem, we propose a keypoint refinement module and integrate it into our framework.


\paragraph{\textbf{Coarse multiview registration}}
We observe that while detector-free methods (such as GeoTransformer) cannot establish continuous Tracks at fine positions, they exhibit semi-dense matches with fixed keypoint locations, which can be directly used in our incremental registration framework. To this end, we build a coarse but complete registration model with Tracks $\{ \mathbb{T}_j \} = \{ \tilde{q}_j, \{ \tilde{p}_i|i=1,...,n \} \}$ based on the semi-dense matches.

\paragraph{\textbf{Fine multiview registration}}
We then introduce a module for the refinement of Track $\mathbb{T}_j$ by adjusting the locations of its keypoints  $\{ \tilde{p}_i|i=1,...,n \}$. Specifically, for each keypoints $\tilde{p}_i$, we first crop a local patch $\mathcal{L}_i\in \mathbb{R} ^{k\times 3}$ around it and extract its point-wise feature $F_i\in \mathbb{R} ^{k\times d}$. Subsequently, we randomly select one patch $\mathcal{L}_r$ as a reference and match it with the other patches $\{ \mathcal{L}_i | i = 1,...,n \And i \neq r \}$ based on the feature similarity. We score each point $p$ in $\mathcal{L}_i$ by summing up its similarity scores of its nearest neighbors in the other patches. The point $p$ with the highest score and its nearest neighbors are chosen as the refined keypoint $\{ \overline{p}_i |i=1,...,n \}$. Finally, the refined tracks  $\{ \mathbb{T}_j \} = \{ \overline{q}_j, \{ \overline{p}_i|i=1,...,n \} \}$ are input into a global BA to optimize the absolute scan poses $\{ (R_i,t_i) \}$ and  3D landmarks $\{ q_j \}$ in a joint manner to get a fine multiview registration result.

\section{Experiments}
\label{sec:experiments}
In this section, we introduce the selection of baseslines in Section~\ref{subsec:baseslines}, and datasets, evaluation metrics in Section~\ref{subsec:datasets_metrics}. Experimental results and ablation studies are reported in Section~\ref{subsec:evalution_results} and Section~\ref{subsec:ablation_studies}, respectively.
\subsection{Baselines}
\label{subsec:baseslines}
We compare our method with several baselines: EIGSE3~\cite{Arrigoni_Rossi_Fusiello_2016}, L1-IRLS ~\cite{Chatterjee_Govindu_2018}, RotAvg~\cite{Chatterjee_Govindu_2018}, LMVR~\cite{gojcic2020learning}, LITS~\cite{yew2021learning}, HARA~\cite{Lee_Civera} and SGHR~\cite{wang2023robust}. All of the above methods regard the multiview registration as a graph optimization problem and follow the common pipeline of pairwise registration and transformation synchronization. Besides, we replicate a registration method~\cite{Huber_Hebert_2003} based on the incremental approach of Minimum Spanning Tree (MST). Our implementation follows the procedure outlined in the original paper, which involves constructing a MST representation of the scans and iteratively updating the transformation along the MST path. 

We follow the same experimental settings as \cite{wang2023robust} and use different pairwise registration, including detector-based methods( FPFH~\cite{Choy_Park_Koltun_2019}, SpinNet~\cite{ao2021spinnet}, YOHO~\cite{wang2022you}) and detector-free method (GeoTransformer~\cite{qin2022geometric}). The first type, denoted as Full, maintains a fully-connected scan graph without pruning any edges. In contrast, the second type, labeled as Sparse, leverages global features to construct a sparse scan graph.

\begin{figure}[t]
  \centering
  \includegraphics[width=\textwidth]{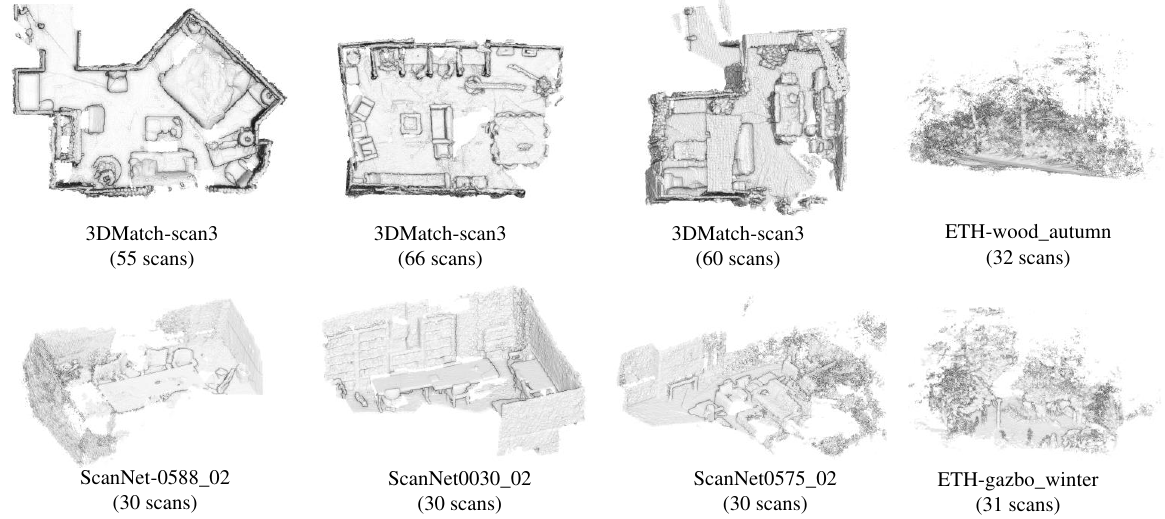}
  \caption{\textbf{Qualitative results on the 3DMatch~\cite{zeng20163dmatch}, ScanNet~\cite{dai2017scannet}, and ETH~\cite{4409092} datasets.}}
  \label{fig:results}
\end{figure}
\subsection{Datasets and Metrics}
\label{subsec:datasets_metrics}
\paragraph{\textbf{Datasets:}}
We evaluate our method on three datasets: 3D(Lo)Match~\cite{zeng20163dmatch}, ScanNet~\cite{dai2017scannet}, and ETH~\cite{4409092}, which cover a variety of indoor and outdoor scenes with different levels of complexity and noise. 3D(Lo)Match contains 62 indoor scenes captured by the LiDAR scanner, we follow \cite{wang2023robust} to test on 8 scenes. Each scene contains an average of 54 scans. The point cloud scans in 3DMatch have over 30\% overlap, while those in 3DLoMatch have low overlap of $10\% \sim 30\%$. ScanNet contains over 1500 RGB-D sequences of indoor scenes, annotated with 3D camera poses and surface reconstructions. ETH contains 4 outdoor scenes captured by a stereo rig mounted on a car and each scene includes an average of 33 scans. We follow the protocol introduced in ~\cite{wang2023robust} and test our method on 44 scenes in total. 
\paragraph{\textbf{Metrics:}}
To compare our results with the baselines, we evaluate the performance with two metrics: 1) Registration Recall (RR). RR is defined as the fraction of scans whose average distance between the points under the estimated transformation $(R_{pre}, t_{pre})$ and these points under ground truth transformation $(R_{gt},  t_{gt})$ are below certain thresholds. 2) Empirical Cumulative Distribution function (ECDF). ECDF describes the function of the error distribution and is defined as the fraction of the scans whose Euclidean distance between $R_{pre}$ and $R_{gt}$ or Euclidean distance between $t_{pre}$ and $t_{gt}$ are below a set series of thresholds. We follow the same protocol as \cite{wang2023robust} to compute these metrics and report the results in the following sections.

\begin{figure}[t]
  \centering
  \includegraphics[width=\textwidth]{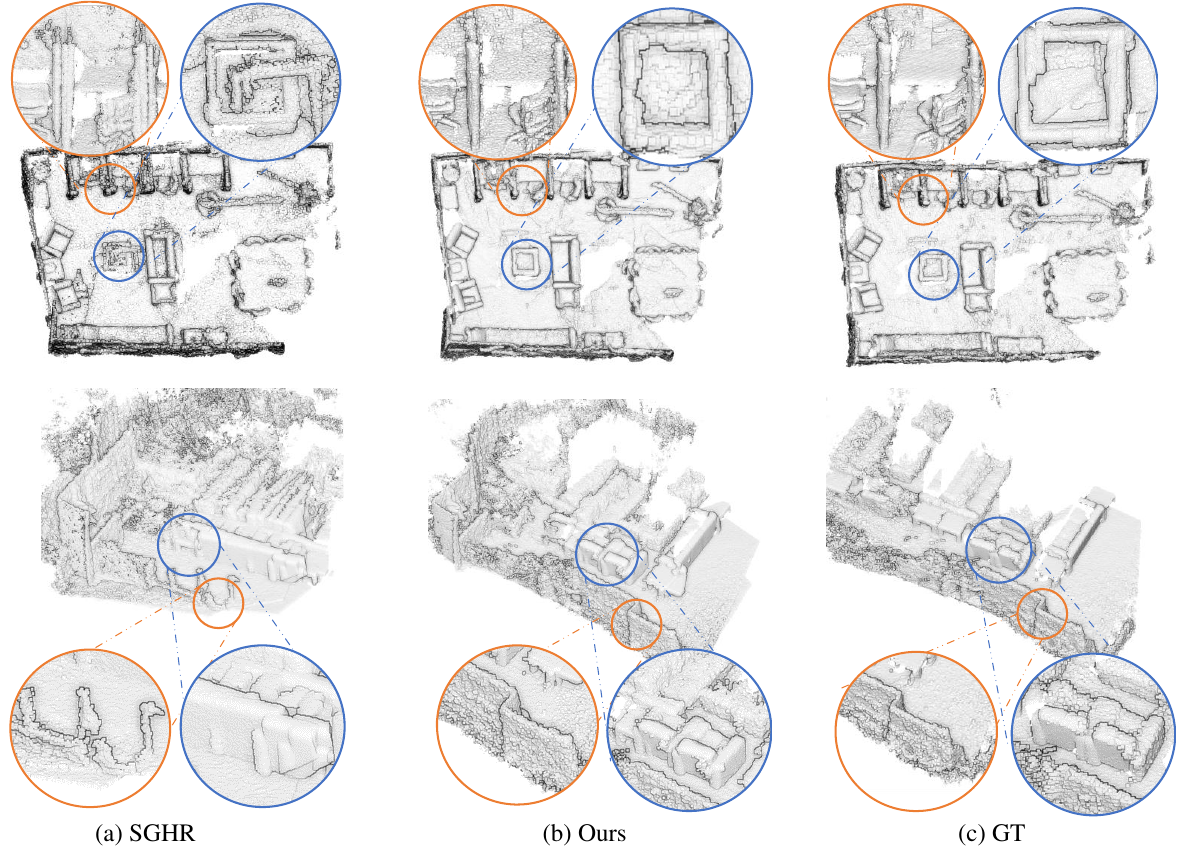}
  \caption{\textbf{Qualitative results.} Our method is compared to SGHR~\cite{wang2023robust}. The findings indicate that we obtain a more accurate geometric structure of the identical region.}
  \label{fig:robust}
\end{figure}

\subsection{Results}
\label{subsec:evalution_results}
We present the superiority of our method in terms of accuracy, robustness, and completeness via detailed experiments. The quantitative results on 3D(Lo)Match, ScanNet, and ETH datasets are presented in \cref{tab:3dmatch}, \cref{tab:scannet} and \cref{tab:eth}, respectively. \figref{fig:results} shows the qualitative registration result of several sampled scenes from the three datasets. 

\begin{table}[t]
    \caption{\textbf{Quantitative results on 3D(Lo)Match dataset.} We present the registration recall on the 3DMatch (referred as "3D") and 3DLoMatch (referred as "3DL") datasets~\cite{zeng20163dmatch}. Our report includes results obtained from various pairwise registration algorithms, namely SpinNet~\cite{ao2021spinnet}, YOHO~\cite{wang2022you}, and GeoTransformer~\cite{qin2022geometric}.}
  \centering
  \resizebox{\textwidth}{!}{
    \begin{tabular}{c|c|c|c|c|c}
    \toprule
    \multicolumn{1}{c|}{\textit{\makecell[c]{Scan\\ Graph}}} & \textit{Method} & \textit{Pair} & \multicolumn{1}{c|}{\makecell[c]{SpinNet~\cite{ao2021spinnet} \\3D/3DL-RR (\%)}} & \multicolumn{1}{c|}{\makecell[c]{YOHO~\cite{wang2022you} \\3D/3DL-RR (\%)}} & \multicolumn{1}{c}{\makecell[c]{GeoTrans. ~\cite{qin2022geometric} \\3D/3DL-RR (\%)}} \\
    \hline
    \multirow{8}{*}{Full} & EIGSE3 ~\cite{Arrigoni_Rossi_Fusiello_2016} & 11905 & 20.8/13.6 & 23.2/6.6 & \multicolumn{1}{c}{17.0/9.1} \\
          & Ll-IRLS~\cite{Chatterjee_Govindu_2018} & 11905 & 49.8 / 29.4 & 52.2/32.2 & \multicolumn{1}{c}{55.7/37.3} \\
          & RotAvg~\cite{Chatterjee_Govindu_2018} & 11905 & 59.3/38.9 & 61.8/44.1 & \multicolumn{1}{c}{68.6/56.5} \\
          & LITS~\cite{yew2021learning} & 11905 & 68.1 /47.9 & 77.0/59.0 & \multicolumn{1}{c}{84.2/73.0} \\
          & HARA~\cite{Lee_Civera} & 11905 & 82.7/63.6 & 83.1 /68.7 & \multicolumn{1}{c}{83.4/68.5} \\
          & SGHR~\cite{wang2023robust} & 11905 & 93.3/77.2 & 93.2/76.8 & \multicolumn{1}{c}{91.5/82.4} \\
\cdashline{2-6}          & MST~\cite{Huber_Hebert_2003} & 11905 & 80.3/52.7 & 80.9/54.3 & \multicolumn{1}{c}{82.1/59.7} \\
          & Ours  & 11905 & \textbf{93.3/78.5} & \textbf{93.9/80.1} & \multicolumn{1}{c}{\textbf{93.7/84.7}} \\
    \hline
    \multirow{3}{*}{Sparse} & SGHR~\cite{wang2023robust} & 2798  & \textbf{94.9} / 80.0 & 96.2/81.6 & \multicolumn{1}{c}{95.9 / 83.0} \\
\cdashline{2-6}          & MST~\cite{Huber_Hebert_2003} & 2798  & 80.8/53.6 & 81.4/55.0 & \multicolumn{1}{c}{82.9/60.1} \\
          & Ours  & 2798  & 94.6/\textbf{82.5} & \textbf{96.3/84.4} & \multicolumn{1}{c}{\textbf{97.3/87.1}} \\
    \bottomrule
    \end{tabular}%
    }
  \label{tab:3dmatch}%
\end{table}%

\begin{table}[t]
    \caption{\textbf{Quantitative results on ScanNet dataset.} We present the registration performance on the ScanNet dataset. For all methods, the pairwise registration algorithm employed is YOHO~\cite{wang2022you}, with the exception of LMVR~\cite{gojcic2020learning}, which incorporates pairwise registration within its pipeline.}
  \centering
  \resizebox{\textwidth}{!}{
    \begin{tabular}{c|c|c|cccccc|cccccc}
    \toprule
    \multirow{2}[4]{*}{\textit{\makecell[c]{Scan\\ Graph}}} & \multirow{2}[4]{*}{\textit{Method}} & \multirow{2}[4]{*}{\textit{Pair}} & \multicolumn{6}{c|}{Rotation Error}           & \multicolumn{6}{c}{Translation Error (m)} \bigstrut\\
\cline{4-15}          &       &       & 3°    & 5°    & 10°   & 30°   & 45°   & Mean/Med & 0.05  & 0.1   & 0.25  & 0.5   & 0.75  & Mean/Med \bigstrut\\
    \hline
    \multirow{10}[4]{*}{Full} & LMVR~\cite{gojcic2020learning} & 13920 & 48.3  & 53.6  & 58.9  & 63.2  & 64.0  & 48.1°/33.7° & 34.5  & 49.1  & 58.5  & 61.6  & 63.9  & 0.83/0.55 \bigstrut[t]\\
          & LITS~\cite{yew2021learning} & 13920 & 47.4  & 58.4  & 70.5  & 78.3  & 79.7  & 27.6°/- & 29.6  & 47.5  & 66.7  & 73.3  & 77.6  & 0.56/- \\
          & EIGSE3~\cite{Arrigoni_Rossi_Fusiello_2016} & 13920 & 19.7  & 24.4  & 32.3  & 49.3  & 56.9  & 53.6°/48.0° & 11.2  & 19.7  & 30.5  & 45.7  & 56.7  & 1.03/0.94 \\
      & Ll-IRLS~\cite{Chatterjee_Govindu_2018} & 13920 & 38.1  & 44.2  & 48.8  & 55.7  & 56.5  & 53.9°/47.1° & 18.5  & 30.4  & 40.7  & 47.8  & 54.4  & 1.14/1.07 \\
          & RotAvg~\cite{Chatterjee_Govindu_2018} & 13920 & 44.1  & 49.8  & 52.8  & 56.5  & 57.3  & 53.1°/44.0° & 28.2  & 40.8  & 48.6  & 51.9  & 56.1  & 1.13/1.05 \\
          & LITS~\cite{yew2021learning} & 13920 & 52.8  & 67.1  & 74.9  & 77.9  & 79.5  & 26.8°/27.9° & 29.4  & 51.1  & 68.9  & 75.0  & 77.0  & 0.68/0.66 \\
          & HARA~\cite{Lee_Civera} & 13920 & 54.9  & 64.3  & 71.3  & 74.1  & 74.2  & 32.1°/29.20 & 35.8  & 54.4  & 66.3  & 69.7  & 72.9  & 0.87/0.75 \\
          & SGHR~\cite{wang2023robust} & 13920 & 57.2  & 68.5  & 75.1  & 78.1  & 78.8  & 26.4°/19.50 & 39.4  & 61.5  & 72.0  & 75.2  & 77.6  & 0.70/0.59 \bigstrut[b]\\
\cdashline{2-15}          & MST~\cite{Huber_Hebert_2003} & 13920 & 28.2  & 42.6  & 54.4  & 70.4  & 73.9  & 34.3°/23.6° & 27.4  & 45.8  & 63.2  & 66.0  & 69.4  & 0.82/0.65 \bigstrut[t]\\
          & Ours  & 13920 & \textbf{61.0}  & \textbf{75.0}  & \textbf{79.5}  & \textbf{85.7}  & \textbf{87.7}  & \textbf{14.7°/11.8°} & \textbf{40.4}  & \textbf{66.8}  & \textbf{83.7}  & \textbf{87.5}  & \textbf{90.0}  & \textbf{0.31/0.14} \bigstrut[b]\\
    \hline
    \multirow{3}[3]{*}{Sparse} & SGHR~\cite{wang2023robust} & 6004  & \textbf{59.1}  & 73.1  & \textbf{80.8}  & 82.5  & 83.0  & 21.7°/19.0° & \textbf{39.9}  & \textbf{64.1}  & \textbf{76.7}  & 79.0  & 81.9  & 0.56/0.49 \bigstrut\\
\cdashline{2-15}          & MST~\cite{Huber_Hebert_2003} & 6004  & 27.3  & 40.3  & 52.5  & 67.1  & 69.5  & 42.0°/37.8° & 26.6  & 45.1  & 60.9  & 64.4  & 67.2  & 0.89/0.76 \bigstrut[t]\\
          & Ours  & 6004  & 58.6  & \textbf{73.4}  & 79.7  & \textbf{83.9}  & \textbf{85.6 } & \textbf{19.8°/15.6°} & 39.6  & 63.9  & 76.3  & \textbf{81.9}  & \textbf{85.4}  & \textbf{0.55/0.37 }\\
          \bottomrule
    \end{tabular}}%
  \label{tab:scannet}%
\end{table}%
\begin{table}[!h]
\caption{\textbf{Quantitative results on ETH dataset.} We report results using different pairwise registration algorithms (FCGF~\cite{Choy_Park_Koltun_2019}, SpinNet~\cite{ao2021spinnet}, YOHO~\cite{wang2022you}).}
  \centering
  \resizebox{\textwidth}{!}{
    \begin{tabular}{c|c|c|ccc}
    \toprule
    \textit{\makecell[c]{Scan\\ Graph}} & \textit{Method} & \textit{Pair} & \multicolumn{1}{p{8.045em}}{\makecell[c]{FCGF[15] \\RR (\%)}} & \multicolumn{1}{p{8.775em}}{\makecell[c]{SpinNet [1] \\RR (\%)}} & \multicolumn{1}{p{8.225em}}{\makecell[c]{YOHO [49] \\RR (\%)}} \bigstrut\\
    \hline
    \multirow{8}[2]{*}{Full} & EIGSE3 ~\cite{Arrigoni_Rossi_Fusiello_2016} & 2123  & 44.8  & 56.3  & 60.9 \bigstrut[t]\\
          & Ll-IRLS~\cite{Chatterjee_Govindu_2018} & 2123  & 60.5  & 73.2  & 77.2 \\
          & RotAvg~\cite{Chatterjee_Govindu_2018} & 2123  & 67.3  & 82.1  & 85.4 \\
          & LITS~\cite{yew2021learning} & 2123  & 26.3  & 36.4  & 34.8 \\
          & HARA~\cite{Lee_Civera} & 2123  & 72.2  & 79.3  & 85.4 \\
          & SGHR~\cite{wang2023robust} & 2123  & 85.7  & 86.3  & 98.8 \\
        \cdashline{2-6}
          & MST~\cite{Huber_Hebert_2003} & 2123  & 64.3  & 77.6  & 90.4 \\
          & Ours  & 2123  & \textbf{87.5} & \textbf{89.6} & \textbf{98.9} \bigstrut[b]\\
    \hline
    \multirow{3}[1]{*}{Sparse} & SGHR~\cite{wang2023robust} & 516   & 97.4  & \textbf{99.8}  & \textbf{99.1} \bigstrut[t]\\
    \cdashline{2-6}
          & MST~\cite{Huber_Hebert_2003} & 516   & 62.3  & 76.8  & 88.2 \\
          & Ours  & 516   & \textbf{97.6} & 98.3 & 98.6 \\
    \bottomrule
    \end{tabular}
    }%
  \label{tab:eth}%
\end{table}%
\paragraph{\textbf{Accuracy}}
Our proposed method demonstrates superior performance in terms of recall across both the 3DMatch and 3DLoMatch datasets under all conditions. 
Besides, we achieve the lowest mean and median error for rotation and translation when evaluated on the ScanNet dataset. 
Furthermore, our method achieves an impressive registration recall rate of nearly 99\% on the ETH dataset. 
Notably, our method exhibits excellent compatibility,  seamlessly integrating with both detector-based and detector-free registration approaches, and consistently outperforms existing techniques. 
\figref{fig:robust} shows a detailed visual comparison between our method and the state-of-the-art global registration method. 

\paragraph{\textbf{Robustness}}
We also exploit multiview point cloud registration upon the full scan graph, as illustrated in the 'Full' part of \cref{tab:3dmatch}, \cref{tab:scannet}, \cref{tab:eth}(with many incorrect connections), and in the 3DLoMatch dataset with a low overlap ratio, as illustrated in \cref{tab:3dmatch}. 
In such scenarios, our method surpasses comparative baselines by a considerable margin, thereby affirming its robustness. We believe the main reason for this is that our incremental method can detect anomalous matching pairs and reconstruct the scene using more precise geometric constraints. Conversely, the global method, being more sensitive to pairwise initialization, may be badly influenced by noisy or complex data, resulting in suboptimal outcomes. 

\begin{figure}[ht]
  \centering
  \includegraphics[width=\textwidth]{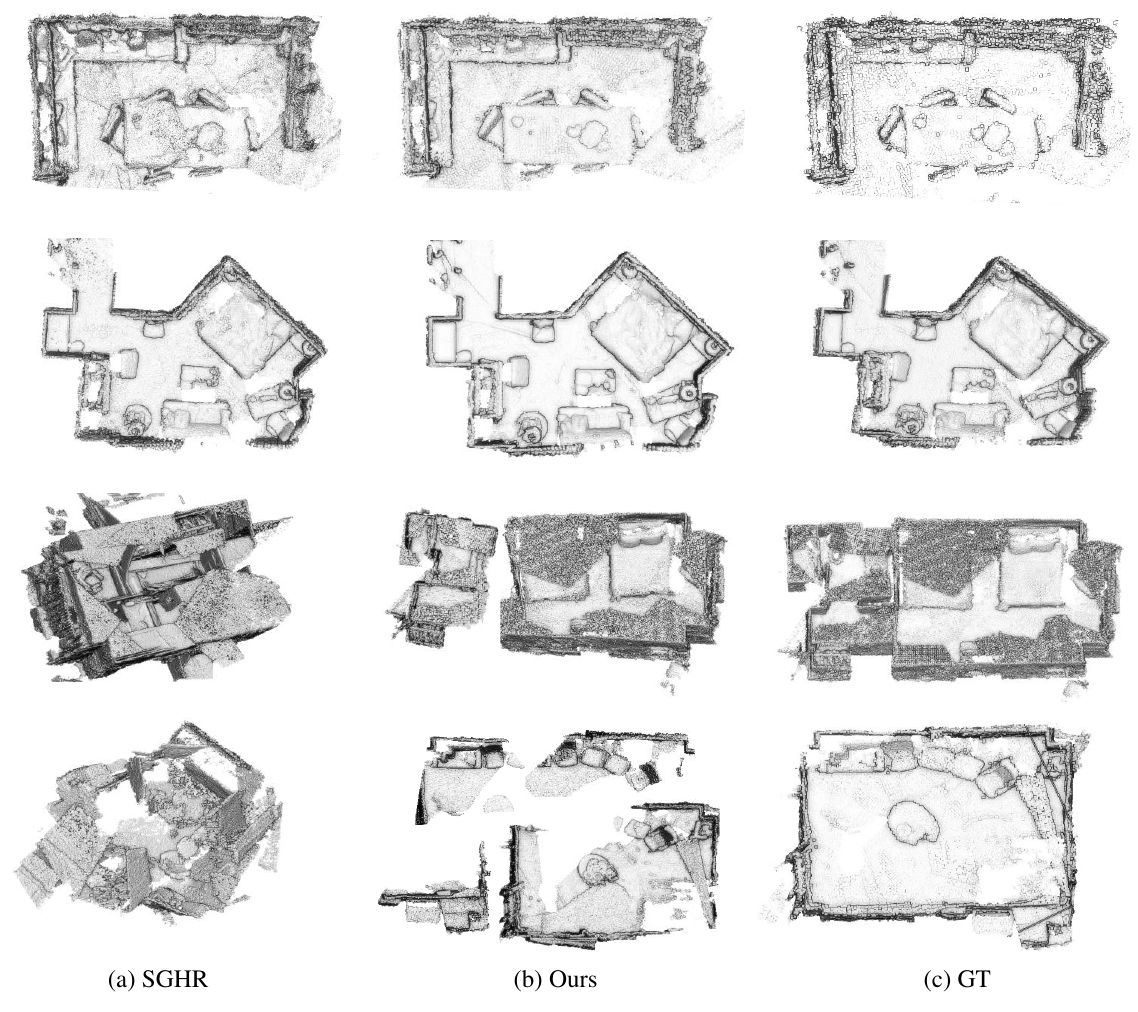}
  \caption{\textbf{Qualitative Comparisons.} The first two rows of cases have higher recall rates(84\%, 73\%), while the last two rows get lower retrieval recall(44\%, 55\%).}
  \label{fig:compelte}
\end{figure}

\paragraph{\textbf{Completeness}}
For scenes with excessive poor pairwise initialization, our incremental processing mechanism prefers to partition the scan graph into multiple sub-connected graphs for independent registration. We report two challenging cases under these conditions, as depicted in the two bottom lines of \figref{fig:compelte}. The global method, which relies on the global consistency principle, yields registration failure. In contrast, our method produces multiple sub-scenes. 
\section{Conclusion}
\label{sec:conclusion}

In this paper, we present a new approach for multiview point cloud registration. In contrast to previous methods that employ global scheme, our framework leverages the recent success of image-based 3D reconstruction to progressively register point clouds into a uniform coordinate system. Our approach consists of three main stages: 1) sparse scan graph generation, 2) incremental scan registration, and 3) Track refinement for detect-free methods. We compare our approach with existing multiview registration methods on three benchmark datasets: 3D(Lo)Match, ScanNet, and ETH, which include a variety of indoor and outdoor scenes, and demonstrated its superior performance in terms of accuracy, robustness and completeness. We believe that our approach can be a valuable tool for various applications that require multiview point cloud registration, such as 3D reconstruction, mapping, and scene understanding.

\bibliographystyle{elsarticle-num}
\bibliography{IncreMVR.bib}
\end{document}